\documentclass{article}

\PassOptionsToPackage{numbers, compress}{natbib}

\usepackage[preprint]{nips_2018}

\usepackage[utf8]{inputenc} 
\usepackage[T1]{fontenc}    
\usepackage{hyperref}       
\usepackage{url}            
\usepackage{booktabs}       
\usepackage{amsfonts}       
\usepackage{nicefrac}       
\usepackage{microtype}      
\usepackage{graphicx}
\usepackage{subfigure}
\usepackage{booktabs} 
\usepackage{amsmath}

\usepackage{algorithmic}
\usepackage{booktabs}       
\usepackage{ amssymb }
\usepackage{float}
\usepackage{bbm}

\DeclareMathOperator*{\relu}{ReLU}

\title{Metalearning with Hebbian Fast Weights}

\author{
  Tsendsuren Munkhdalai \\
  Microsoft Research\\
  Montr\'{e}al, Qu\'{e}bec, Canada\\
  \texttt{tsendsuren.munkhdalai@microsoft.com} \\
  \And
  Adam Trischler \\
  Microsoft Research\\
  Montr\'{e}al, Qu\'{e}bec, Canada\\
  \texttt{adam.trischler@microsoft.com} \\
}

\begin{document}

\maketitle

\begin{abstract}
We unify recent neural approaches to one-shot learning with older ideas of associative memory in a model for metalearning.
Our model learns jointly to represent data and to bind class labels to representations in a single shot.
It builds representations via slow weights, learned across tasks through SGD,
while fast weights constructed by a Hebbian learning rule implement one-shot binding for each new task.
On the Omniglot, Mini-ImageNet, and Penn Treebank one-shot learning benchmarks,
our model achieves state-of-the-art results.
\end{abstract}

\section{Introduction}


There is growing interest in building more flexible, adaptive neural models, particularly within the framework of \emph{metalearning} (learning to learn)~\citep{mitchell1993explanation,andrychowicz2016learning,vinyals2016matching,pmlr-v70-bachman17a}.
The goal of metalearning algorithms is the ability to learn new tasks efficiently, given little training data for each individual task.

In this work we develop a model for metalearning that learns to manipulate a fast, non-parametric rule for binding labels to representations.
Our model unifies ideas from several recent approaches, particularly MetaNets~\cite{pmlr-v70-munkhdalai17a} and Matching Networks~\cite{vinyals2016matching}, with older ideas of associative memory~\cite{kohonen1984}.
Like MetaNets, we use both \emph{fast} and \emph{slow} weights in parallel layers.
The model's slow weights learn deep representations of data.
Its fast weights implement a Hebbian-learning-based associative memory~\cite{hebb2005organization,hopfield1982,kohonen1984}, which binds class labels to representations in a single shot.
The memory operates similarly to the attention kernel over labels of Matching Networks (and the similar key-value memory model of~\citet{weston:15}).
Our model's fast and slow components co-adapt through joint, end-to-end training.

We evaluate our model in the setting of one-shot supervised learning,
wherein a learner is presented with a sequence of classification tasks.
There are two key challenges here.
First, the class definitions, and therefore the labels, vary across tasks.
Second, the training data for each task consists of only a single labeled example per class.
As a consequence, successful models require a capacity for rapid binding in order to memorize new classes on the fly.

Our model operates in two phases: a description phase and a prediction phase.
For each task $\tau \sim p(\tau)$, the model has access to a description $D_\tau$.
In the present setting, this is a set of example datapoints and their corresponding labels: $D_\tau = \{ (x'_i,y'_i) \}^n_{i=1}$.
In the description phase, the model processes $D_\tau$ and binds representations of the inputs $x'_i$ to their labels $y'_i$ in its associative memory.
In the prediction phase, the model predicts labels $y_j$ for unseen datapoints $x_j \in \tau$.
For this it uses its internal representation of $x_j$ to retrieve a bound label from memory.
Retrieving the correct class label depends on building good representations, which should capture underlying features that are agnostic to label permutation.

We demonstrate experimentally that our model achieves state of the art performance on the Omniglot, Mini-ImageNet, and Penn Treebank one-shot learning benchmarks.
In addition to better performance, the simple Hebbian mechanism we adopt for building the fast weights yields significant speedups (up to 100x) over previous metalearning models.

\section{Background}
\label{sec:background}

\subsection{An Inductive Bias for Binding}
The one-shot learning problem can be decomposed into two subproblems: learning good representations of inputs and binding those representations to labels~\cite{vinyals2016matching}.
Representations should encode meaningful, semantic properties of raw inputs.
We expect (some of) these properties to be shared across tasks, although their relative importance to the task-at-hand might vary significantly.
Conversely -- indeed, by definition in one-shot learning -- the label associated to a particular input class does vary from one task to the next and could be considered a proxy for which underlying properties are relevant.
This observation motivates the use of task-specific fast weights to address the binding subproblem, and cross-task slow weights for representations.
We aim to build into our model's fast-weight component an inductive bias for binding.

What is a good way to bind labels rapidly to representations?
Many previous works have answered this question with ``a memory mechanism'' \cite{santoro2016meta,pmlr-v70-munkhdalai17a,mishra2017meta,munkhdalai2017learning}, and we take the same approach here.
Specifically, we use a matrix of fast weights to bind labels (values) to representations (keys).
The fast-weight matrix acts as an associative memory~\cite{kohonen1984,hopfield1982} that we build in one shot, using a simple Hebbian learning rule~\cite{hebb2005organization}.
Based on earlier ideas from neuroscience, this associative model operates similarly to the key-value memory introduced in~\citet{weston:15} and the kernel-based binding approach of Matching Networks~\cite{vinyals2016matching}.
Combined with slow weights, it leads to strong, computationally efficient performance in the few-shot learning setting.

\subsection{Hebbian Learning and Associative Memory}
Hebbian learning~\cite{hebb2005organization} and later extensions like spike-timing-dependent plasticity~\cite{song2000competitive} posit that the synaptic connection between two neurons changes depending on the co-occurrence of pre- and post-synaptic activity.
In particular, the magnitude of a synaptic connection's modification is proportional to the product of pre- and post-activation values.
We can write the Hebbian learning rule for artificial neural networks, in which neurons are neatly organized into layers, as
\begin{equation}
    \label{eq:hebb}
    M_\ell \leftarrow M_\ell + \eta h_{\ell-1}h_\ell^T,
\end{equation}
where $\eta$ is the learning rate, $h_\ell \in \mathbb{R}^{d_\ell}, h_{\ell-1} \in \mathbb{R}^{d_{\ell-1}}$ are the activation vectors of the $\ell$th (i.e., post-synaptic) and $(\ell-1)$th (i.e., pre-synaptic) layer, respectively, and $M_\ell \in \mathbb{R}^{d_{\ell-1} \times d_\ell}$ is the weight matrix that connects the layers.


Long ago, in the early days of neural-network research, there emerged several models of computation and pattern completion called associative memories~\cite{hopfield1982,kohonen1984,kosko1988} that used this learning rule.
One of the first examples was the linear associative memory (LAM) of~\citet{kohonen1984}.
We adopt it for metalearning. A LAM stores key-value vector pairs in a memory matrix $M$ according to a Hebbian outer-product rule.\footnote{The \emph{optimal} LAM uses the Moore-Penrose pseudoinverse in the product rather than the transpose.}
Assuming $n$ key-value vector pairs $(k_i, v_i)$ are to be stored,
\begin{equation}
    \label{eq:lam}
    M = \sum_{i=1}^n k_i v_i^T,
\end{equation}
which we recognize as equivalent to Eq.~\ref{eq:hebb} in the case that the learning rate $\eta=1$.
We ``read'' from a LAM by matrix multiplication with a key vector.
Using $r$ to denote the read result and $q$ to denote the query key,
\begin{align}
    \nonumber
    r &= M^T q \\
      \nonumber
      &= \sum_i v_i k_i^T q \\
      \label{eq:lam-read}
      &= (k_i^T q) v_i + \sum_{j \neq i} (k_j^T q) v_j.
\end{align}
In the ideal case, the query key is approximately equal to one of the stored keys, $k_i$, and these in turn are mutually orthogonal.
When these conditions hold exactly, we recover the key's bound value vector (up to a scalar multiple given by the square of the key's norm, $||k_i||^2$).



Eq.~\ref{eq:lam-read} reveals the LAM's functional similarity to much more recent models: in particular, the key-value storage with soft-attention readout~\cite{bahdanau:15} of Memory Networks~\cite{weston:15} and
the attention kernel over labels in Matching Networks~\cite{vinyals2016matching}.
The readouts of all three mechanisms return a linear combination of stored vectors, conditioned on inner products.
The latter two mechanisms force the combination to be convex with a softmax, which increases interdependence of the inner-product-based coefficients.
An appealing aspect of LAM and other associative memory models is the theory characterizing their storage capacity (see, e.g.,~\cite{kohonen1984}), which could be leveraged in future work.

\section{Model}
Our model operates in two phases: a description phase, wherein it processes the task description $D_{\tau} = \lbrace ( x'_i,y'_i ) \rbrace^n_{i=1}$, and a prediction phase, wherein it acts on unseen datapoints $x_j$ to predict their labels $y_j$.
In an episode of training or test, we sample a task from $p(\tau)$.
The model then ingests the task description and updates its fast weights to bind the observed labels to corresponding input representations.
At prediction time, the model uses its fast weights to make label predictions on unseen task data. We investigate two methods for learning fast weights.

\begin{figure*}[ht]
\begin{center}
\includegraphics[width=1.0\textwidth]{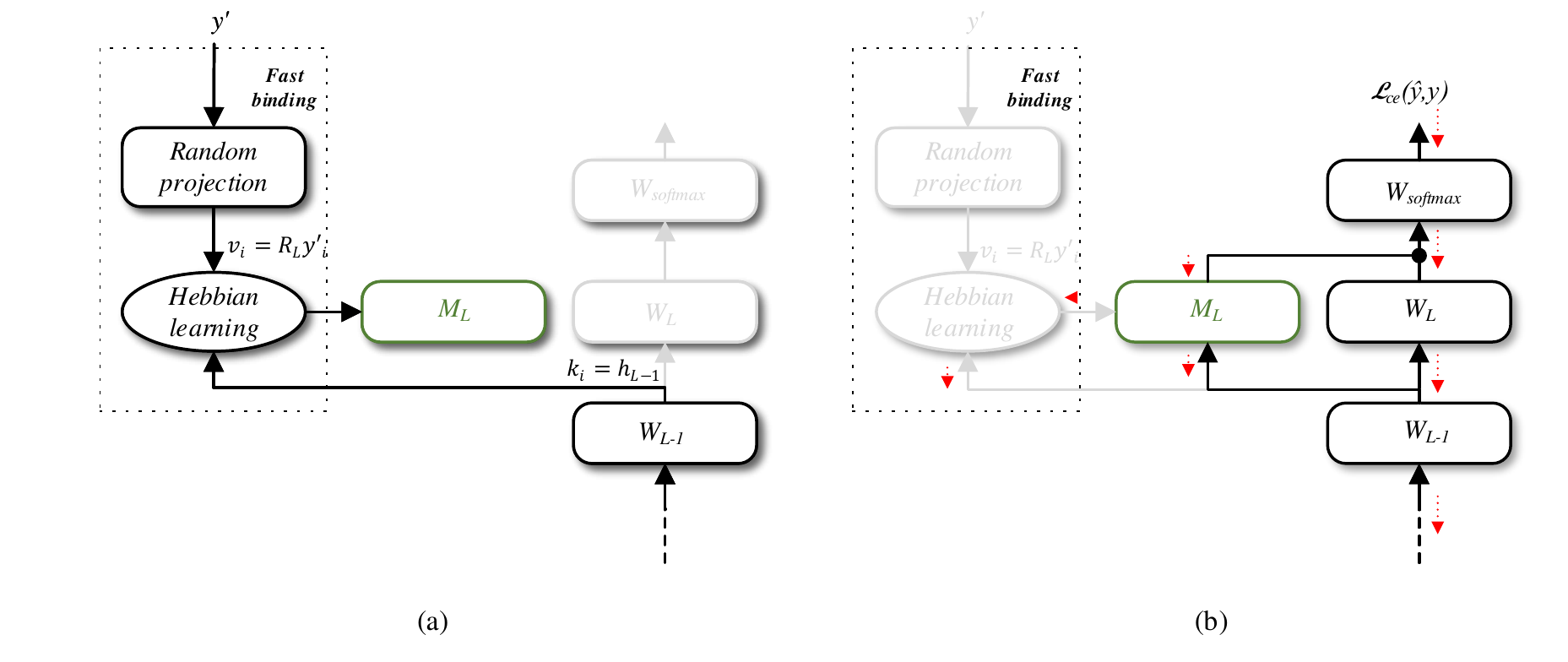} 
\caption{Computation graph of our model operating in description (a) and prediction (b) phases. Inactive components are colored in gray. In the description phase, layers preceding the fast weight-layer $L$ form representations used to construct fast weights in layer $L$, which memorize random projections of one-shot labels $\hat{y}$. In the prediction phase, the random projections of one-shot labels are recalled from the fast-weight memory and fed forward to the softmax layer, whose parameters are optimized to invert the random label projection.
During training, the task error backpropagates through the slow and fast weights as well as the local learning rule, as indicated by the red arrows.
}
\label{fig:fastw}
\end{center}
\end{figure*}

\subsection{Overall Structure}
We construct our model by augmenting with fast weights the pre-softmax layer of a deep neural network.
The computation graph of the model's top two layers is depicted in Figure \ref{fig:fastw}.
We use the layer augmentation schema from MetaNet to define both fast and slow weight components for the penultimate layer, $L$.
The components operate in parallel on layer $L$'s input, producing two activation vectors that we aggregate via element-wise nonlinearity and summation.
Formally, the internal representation $h_L$ at layer $L$ is
\begin{equation}
    h_L = \sigma\left(W_L^T h_{L-1}\right) + \sigma\left(M_L^T h_{L-1}\right)
\end{equation}
for feedforward-style slow weights, where $W_L \in \mathbb{R}^{d_{L-1} \times d_L}$ is the slow-weight matrix, $M_L \in \mathbb{R}^{d_{L-1} \times d_L}$ is fast-weight matrix, and $\sigma$ is the nonlinear activation function.
Layers $L-1$ and below are standard ``slow'' layers without fast-weight augmentation.

\subsection{Hebbian Learning of Fast Weights}
The first mechanism we explore for learning the fast weights is Hebbian learning via the outer-product rule (Eq.~\ref{eq:hebb}).
At each step of the description phase, the model receives an input-label pair, $(x'_i,y'_i)$.
The model builds up an internal representation of the input that it uses as a key, $k_i = h_{L-1} = f(x'_i)$.
It binds this key to the corresponding label representation, $v_i$, constructing $M_L$ according to Eq.~\ref{eq:lam}.
At the top of the network in the softmax, we can simply use the label's one-hot vector for the value, $v_i = y'_i$. This is analogous to the approach used in Matching Networks.
In lower layers like $L$, however, this would lead to a dimension mismatch.
We therefore use what we call \emph{pseudo}values, which are embeddings of the labels into a real-valued vector space.
This gives us flexibility to choose the embedded dimension.
We generate pseudovalues by random projection of the label one-hots, $v_i = R_{L}y'_i$, where $R_L \in \mathbb{R}^{C \times d_L}$ is the projection matrix.
Random projections have useful properties under the right conditions, such as approximate preservation of distances between points (and their inner products if an orthogonal projection is used).
Note also that random label projection was used previously in training neural networks with random feedback~\cite{lillicrap2016random}.
Once the description has been processed, we have a fast-weight matrix $M_L$ that can be used at prediction time to retrieve label (pseudo)values from input representation keys.

\subsection{Mapping Error Gradients to Fast Weights}
An alternative method is to directly map error gradients of one-shot examples to fast weights. This approach is inspired by MetaNet \cite{pmlr-v70-munkhdalai17a} and simplifies that model's procedure for building fast weights. We denote the error gradient with respect to slow weights for each input example as $\nabla_{W_L,i}\mathcal{L}$, which is obtained by passing forward the one-shot examples $(x'_i, y'_i)$ and then backpropagating the task error. We learn a function $g$, parameterized by a feedforward neural network, that takes in the error gradient and outputs the fast weights:
\begin{equation}
\label{eq:stdp_fast2}
    M_L \leftarrow g\left(\sum_{i=1}^{n} \nabla_{W_L,i}\mathcal{L}\right).
\end{equation}

\subsection{Representation Learning via Slow Weights}
Having offloaded the binding task to the fast-weight matrix, we turn to the task of learning good internal representations.
Representations should model, and facilitate discrimination of, the data according to its underlying features.
They should be adapted both to the data and to the binding mechanism.
With these desiderata in mind, we use a deep neural network of ``slow weights'' for representation learning.
Depending on the input data, this network could be recurrent, convolutional, residual, etc. We test all three architectures in our experiments.

During training across all tasks, we optimize through the full network of slow weights \emph{and} the learning rule Eq.~\ref{eq:hebb} or~\ref{eq:stdp_fast2}.
Thus, the slow weights learn not only deep, cross-task representations of the inputs, but also how to manipulate the fast weights via the network's internal states.

\subsection{Training}
\label{sec:fw_con}
We train and test our one-shot-learning model in episodes.
In each episode, we sample a training or test task from $p(\tau)$, process its description $D_\tau$, and then feed its unseen data forward to obtain their label predictions.
Training and test tasks are both drawn from the same distribution, but crucially, we partition the data such that the classes seen at training time do not overlap with those seen at test time.

Over a collection of training episodes, we optimize the model parameters end-to-end via stochastic gradient descent (SGD). Gradients are taken with respect to the (cross-entropy) task losses, $\mathcal{L}_\tau = \sum_j \mathcal{L}_\text{CE}(\hat{y}, y)$.
In this scheme, the model's computation graph contains the operations for updating the fast weights.


\section{Related Work}
\label{related work}

Among the many problems in supervised, reinforcement, and unsupervised learning that can be framed as metalearning, one- and few-shot learning has emerged as a simple and popular test bed.
Few-shot learning problems were previously addressed using metric learning methods \cite{koch2015siamese}. Recently, there has been a shift towards building flexible models for these problems within the learning-to-learn paradigm \cite{mishra2017meta,santoro2016meta}. \citet{vinyals2016matching} unified the training and testing of a one-shot learner under the same procedure and developed an end-to-end, differentiable nearest-neighbor method for one-shot learning. More recently, one-shot optimizers were proposed by \citet{Sachin2017,finn2017model}. The MAML framework~\citep{finn2017model} learns a parameter initialization from which a model can be adapted rapidly to a given task using only a few steps of gradient updates. To learn this initialization MAML uses second-order gradient information. Here we harness only first-order gradient information and the simple Hebbian learning rule.

The MetaNet modifies synaptic connections (weights) between neurons using fast weights \cite{schmidhuber1987,hinton1987using} to implement rapid adaptation.
In this work, we focus on Hebbian-learning based fast weights, comparing them to MetaNet-style fast weights based on gradient information.
Other previous work on metalearning has also formulated the problem as two-level learning: specifically, slow learning of a meta model across several tasks, and fast learning of a base model that acts within each task \cite{schmidhuber1987,bengio1990learning,hochreiter2001learning,mitchell1993explanation,vilalta2002perspective,mishra2017meta}.
\citet{schmidhuber1993self} discussed the use of network weight matrices themselves for continuous adaptation in dynamic environments.

As this work was ongoing,~\citet{rae2018fast} proposed the Hebbian Softmax layer, whose parameters are modified by an interpolation between Hebbian-learning and SGD updates.
As in our case, the authors' motivation was to achieve fast binding for rarer classes, especially in the language-modelling setting where they show strong results.
Our model uses Hebbian learning to build fast weights for augmented layers preceding the softmax, rather than to update the softmax parameters themselves. With a focus on meta and one-shot learning, our goal is to learn to control a local learning rule for fast weights via slow weights and a slow optimization procedure (SGD).


\section{Experimental Evaluation}
\label{sec_results}
\begin{table*}[t] 
  \caption{Omniglot few-shot classification test accuracy for fast weights learned with error gradient ($\nabla$) and Hebbian learning (Hebb).}
  \label{tab:omni}
  \small
  \centering
  \begin{tabular}{lcccc}
    \toprule
    {} & \multicolumn{2}{c}{\bf 5-way} & \multicolumn{2}{c}{\bf 20-way}   \\
    \cmidrule(l{3pt}r{3pt}){2-3} \cmidrule(l{3pt}r{3pt}){4-5}
    \bf Model & \bf 1-shot & \bf 5-shot & \bf 1-shot & \bf 5-shot             \\
    \midrule
    Siamese Net \cite{koch2015siamese} & 97.3 & 98.4 & 88.2 & 97.0 \\
    MANN \cite{santoro2016meta} & 82.8 & 94.9 & - & - \\
    Matching Nets \cite{vinyals2016matching} & 98.1 & 98.9 & 93.8 & 98.5 \\
    MAML \cite{finn2017model} & 98.7 $\pm$ 0.4 & \bf 99.9 $\pm$ 0.3 & 95.8 $\pm$ 0.3 & 98.9 $\pm$ 0.2 \\
    MetaNet \cite{pmlr-v70-munkhdalai17a} & 98.95 & - & 97.0 & - \\
    TCML \cite{mishra2017meta} & 98.96 $\pm$ 0.2 & 99.75 $\pm$ 0.11 & \bf 97.64 $\pm$ 0.3 & \bf 99.36 $\pm$ 0.18 \\
    adaCNN (DF) \cite{munkhdalai2017learning} & 98.42 $\pm$ 0.21 & 99.37 $\pm$ 0.28 & 96.12 $\pm$ 0.31 & 98.43 $\pm$ 0.05 \\
    \midrule
    fwCNN ($\nabla$) & 98.3 $\pm$ 0.16 & 99.32 $\pm$ 0.32 & 94.98 $\pm$ 0.3 & 97.47 $\pm$ 0.13 \\
    fwCNN (Hebb) & \bf 99.4 $\pm$ 0.13 & 99.77 $\pm$ 0.1 & \bf 97.27 $\pm$ 0.11 & 98.85 $\pm$ 0.05 \\
    
    \bottomrule
  \end{tabular}
\end{table*}
\begin{table*}[t] 
\caption{Mini-ImageNet few-shot classification test accuracy for fast weights learned with error gradient ($\nabla$) and Hebbian learning (Hebb).}
  \label{tab:mini}
  \small
  \centering
  \begin{tabular}{lcc}
    \toprule
    {} & \multicolumn{2}{c}{\bf 5-way} \\
    \cmidrule(l{3pt}r{3pt}){2-3} 
    \bf Model & \bf 1-shot & \bf 5-shot \\
    \midrule
    Matching Nets \cite{vinyals2016matching} & 43.6 & 55.3 \\
    MetaLearner LSTM \cite{Sachin2017} & 43.4 $\pm$ 0.77 & 60.2 $\pm$ 0.71 \\
    MAML \cite{finn2017model} & 48.7 $\pm$ 1.84 & 63.1 $\pm$ 0.92 \\
    MetaNet \cite{pmlr-v70-munkhdalai17a} &  49.21 $\pm$ 0.96 & - \\
    adaCNN (DF) & 48.34 $\pm$ 0.68 & 62.00 $\pm$ 0.55 \\
    \midrule
    fwCNN ($\nabla$) & 47.68 $\pm$ 0.63 & 59.48 $\pm$ 0.25 \\
    fwCNN (Hebb) & \bf 50.21 $\pm$ 0.37 & \bf 64.75 $\pm$ 0.49 \\
    \midrule \midrule
    TCML \cite{mishra2017meta} & 55.71 $\pm$ 0.99 & 68.88 $\pm$ 0.92 \\
    adaResNet (DF) \cite{munkhdalai2017learning} & \bf 56.88 $\pm$ 0.62 & \bf 71.94 $\pm$ 0.57 \\
    \midrule
    fwResNet ($\nabla$) & 54.14 $\pm$ 0.32 & 66.77 $\pm$ 0.7 \\
    fwResNet (Hebb) & \bf 56.84 $\pm$ 0.52 & 71.00 $\pm$ 0.34 \\
    \bottomrule
  \end{tabular}
\end{table*}
\begin{table*}[t] 
\caption{Penn Treebank few-shot classification test accuracy for fast weights learned with error gradient ($\nabla$) and Hebbian learning (Hebb).}
  \label{tab:lm}
  \small
  \centering
  \begin{tabular}{lccc}
    \toprule
    {} & \multicolumn{3}{c}{\bf 5-way (400 random/all-inclusive)} \\
    \cmidrule(l{3pt}r{3pt}){2-4} 
    \bf Model & \bf 1-shot & \bf 2-shot & \bf 3-shot \\
    \midrule
    LSTM-LM oracle \cite{vinyals2016matching} & 72.8 & 72.8 & 72.8 \\
    Matching Nets \cite{vinyals2016matching} & 32.4 & 36.1 & 38.2 \\
    
     2-layer adaLSTM ($\nabla$) \cite{munkhdalai2017learning} & \bf 43.1/43.0 & \bf	52.05/54.2 & \bf	57.35/58.4 \\
     2-layer LSTM + adaFFN (DF) \cite{munkhdalai2017learning} & 33.65/35.3 &	46.6/47.8 &	51.4/52.6 \\
     2-layer adaLSTM (DF) \cite{munkhdalai2017learning} & 41.25/43.2 & \bf	52.1/52.9 & \bf	57.8/58.8 \\
    \midrule
     2-layer LSTM + fwMLP ($\nabla$) & 35.5/36.1 &	45.5/46.9 &	49.2/52.7 \\
     2-layer LSTM + fwMLP (Hebb) & \bf 43.75/44.5 &	47.65/48.8 &	51.6/52.8 \\
     
    \bottomrule
  \end{tabular}
\end{table*}
We evaluate the proposed fast weight learning mechanisms on one-shot and few-shot tasks from the vision and language domains.
Below we describe the datasets we evaluate on and the according preprocessing steps, followed by test results. We run all experiments five times and report average performance.

\subsection{One-shot Image Classification}
We used two widely adopted one-shot classification benchmarks from the vision domain: the Omniglot and Mini-ImageNet datasets.
{\bf Omniglot} consists of images from 1623 classes from 50 different alphabets, with only 20 images per class \cite{lake2015human}. As in previous studies, we randomly selected 1200 classes for training and 423 for testing and augmented the training set with 90, 180 and 270 degree rotations. We resized the images to $28 \times 28$ pixels for computational efficiency.

For the Omniglot benchmark we performed 5- and 20-way classification tests, each with one or five labeled examples from each class as the description $D_\tau$. We use a convolutional network (CNN) with 64 filters. This network has 5 convolutional layers, each of which uses $3 \times 3$ convolutions followed by the leaky ReLU nonlinearity and a $2 \times 2$ max-pooling layer. Convolutional layers are followed by a fully connected (FC) layer with leaky ReLU and then a softmax output layer. We learn fast weights for the FC layer, referring to this model as ``fwCNN.''
Full implementation details can be found in Appendix~\ref{sec:imp_details}.
As Table \ref{tab:omni} shows, our fwCNN (Hebb) model achieves state-of-the-art results on the Omniglot 1-shot tasks.

{\bf Mini-ImageNet} is a set of $84 \times 84$-pixel color images from 100 classes (64/16/20 for training/validation/test splits) and each class has 600 exemplar images. We ran our experiments on the class subset released by \citet{Sachin2017}.
Compared to Omniglot, Mini-ImageNet has fewer classes (100 vs 1623) with more labeled examples for each class (600 vs 20). Given this larger number of examples,  we evaluated a similar fwCNN model with 32 filters, as well as a model with more sophisticated ResNet components (``fwResNet''), on the Mini-ImageNet 5-way classification tasks. The ResNet architecture follows that of TCML~\citep{mishra2017meta} with the following exceptions. Instead of two $1 \times 1$ convolutional layers with 2048 and 512 filters we use only a single such layer with 1024 filters. Convolutional layers are followed by a FC layer with 1024 units and leaky ReLU activation, and then a softmax layer. We incorporate fast weights into the final FC layer as above. We observed that adding uniform noise to the one-hot label representations before random projection slightly improves the performance of fwCNN on this task.
Full implementation details can be found in Appendix~\ref{sec:imp_details}.

For every 400 training tasks, we tested the model for another 400 tasks sampled from the validation set. If the model performance exceeded the previous best validation result, we applied it to the test set.

Unlike Omniglot, there remains significant room for improvement on Mini-ImageNet.
As shown in Table \ref{tab:mini}, on this more challenging task, CNN-based models with fast weights improve over the previous best CNN-based approaches (MAML and MetaNet) by 1-1.6\%. The more sophisticated fwResNet model achieves competitive performance with the state-of-the-art adaResNet model~\cite{munkhdalai2017learning}.


\subsection{Few-shot Language Modeling}

To evaluate the effectiveness of recurrent models augmented with fast weights, we ran experiments on the few-shot \textbf{Penn Treebank} (PTB) language modeling task introduced by \citet{vinyals2016matching}.
In this task, a model is given a query sentence with one missing word and a support set (i.e., description) of one-hot-labeled sentences that also have one missing word each. One of the missing words in the description set is identical to that missing from the query sentence.
Following \citet{vinyals2016matching}, we preprocessed and split the PTB sentences into training and test such that, for the test set, target words for prediction and the sentences in which they appear are unseen during training.

We stack a fully-connected layer augmented with fast weights plus an output softmax layer on top of a 2-layer LSTM.
We call this ``LSTM+fwMLP''.
Only the model's FC layer can adapt to the task via fast weights. The LSTM encoder builds up the context for each word to provide a non-task-specific representation to the output module.
We represent words with randomly initialized embeddings.
Full implementation details can be found in Appendix~\ref{sec:imp_details}.

We used two different methods to form test tasks. First, we randomly sampled 400 tasks from the test data and report the average accuracy. Second, we make sure to include all test words in the task formulation. We randomly partition the 1000 target words into 200 groups and solve each group as a task. In the random approach there is a chance that a word could be missed or included multiple times in different tasks.

Table \ref{tab:lm} summarizes our results. The approximate upper bound achieved by the oracle LSTM-LM of \citet{vinyals2016matching} is 72.8\%. The current best accuracy -- around 58\% on the 3-shot task -- is achieved by a 2-layer adaLSTM which learns to adapt its hidden state via conditionally shifted neurons \cite{munkhdalai2017learning}.
Our LSTM + fwMLP (Hebb) model achieves a slightly better result than the adaLSTM model on the one-shot task. Providing more sentences for the target word increases performance, as expected, but not as much as the adaLSTM. The adaLSTM model may have more capacity to exploit the additional description sentences through its adaptation of LSTM hidden states.

On all benchmarks, the fast-weight models with Hebbian learning improve over those that learn their fast weights through gradients. The latter type is a direct simplification of MetaNet, and underperforms that model also.

\subsection{Model Ablation and Timing}

\begin{figure}[h] 
\begin{center}
\centerline{\includegraphics[width=0.45\textwidth, trim={2cm 2.5cm 2.2cm 2cm},clip]{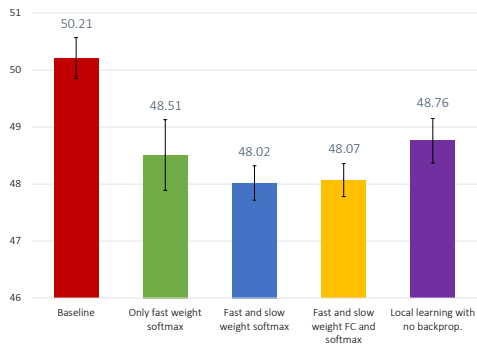}}
\caption{Model ablation for fwCNN tested on the Mini-ImageNet one-shot task. Comparisons between model variants are averaged over 5 independent seeds. Red: the baseline, FC layer $L$ has both fast and slow weights; Green: the softmax layer uses only fast weights; Blue: both fast and slow weights in the softmax layer only; Yellow: Both FC and softmax layers have fast and slow weights; Violet: We truncate task-loss gradients in the fast weights. }
\label{fig:abl_mini}
\end{center}
 \vskip -.2in
\end{figure}

\begin{figure}[ht] 
\begin{center}

\centerline{\
\includegraphics[width=0.45\textwidth, trim={2cm 2.5cm 2.2cm 2cm},clip]{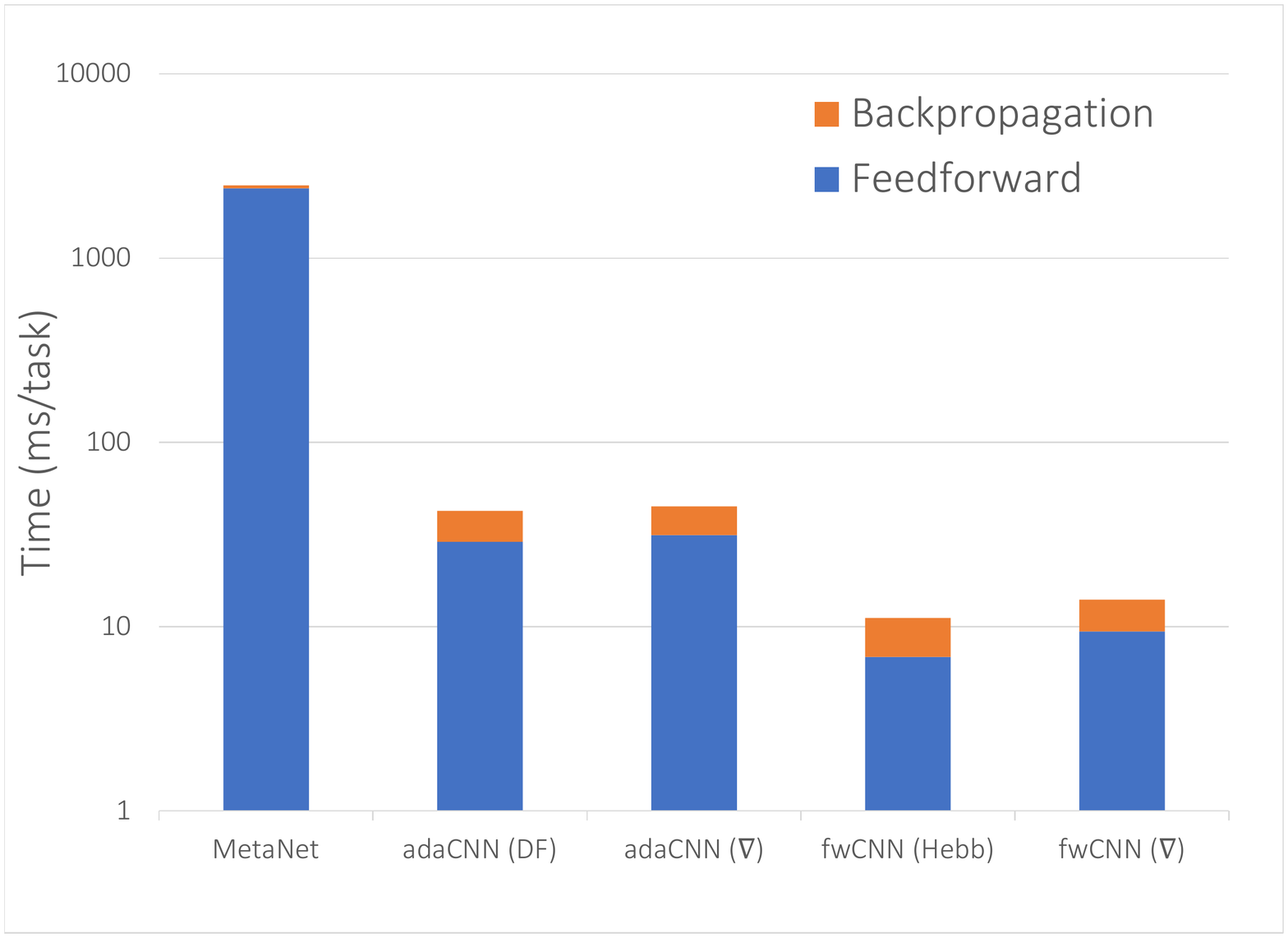}
\includegraphics[width=0.45\textwidth, trim={2cm 2.5cm 2.2cm 2cm},clip]{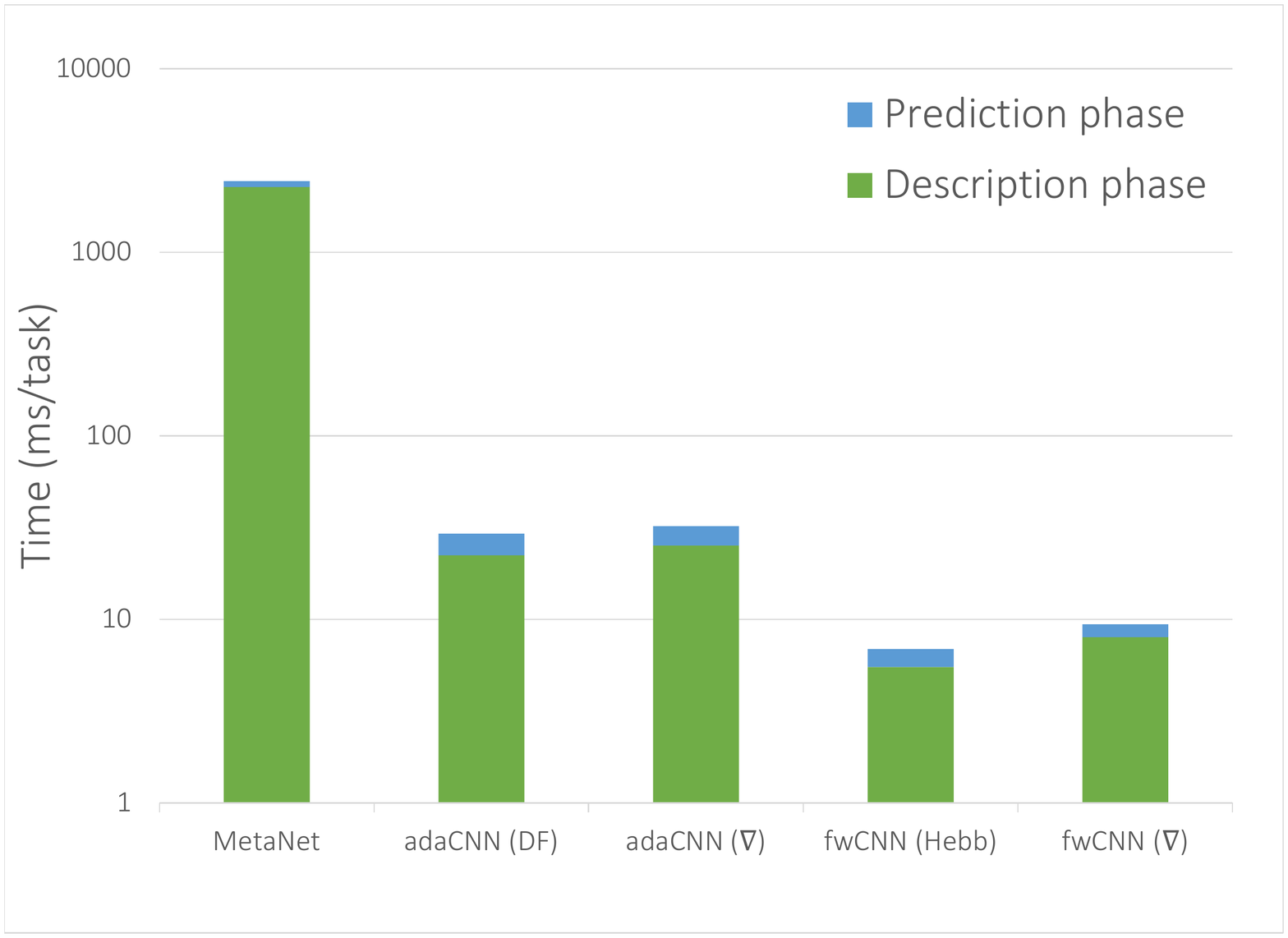}
}
\caption{Training (left) and inference (right) speeds of MetaNet, adaCNN variants, and fwCNN variants are compared on the 1-shot, 5-way Mini-ImageNet task. 
The y-axis shows wall-clock time (ms/task) in log scale. Training time includes the feedforward computations and the parameter updates. Inference time includes computations for the description and prediction phases.}
\label{fig:timing_mini}
\end{center}
 \vskip -.2in
\end{figure}

To better understand our model, we performed an ablation study of fwCNN (Hebb) trained on Mini-ImageNet. We also compared wall-clock timing of the MetaNet, adaCNN ($\nabla$), adaCNN (DF), fwCNN ($\nabla$) and fwCNN (Hebb) models. Results are shown in Figure \ref{fig:abl_mini} and Figure \ref{fig:timing_mini}.

We hypothesized that incorporating fast weights into layers below the softmax layer would increase model capacity, enabling later layers to learn to adjust the fast-weight memory readout (which may be ``blurred'' or spurious).
To test this, we placed fast weights in the softmax layer instead of immediately below.
As indicated by the blue bar, this led to an accuracy drop of 2\%.
We also investigated the effect of imposing a fast-weight ``bottle-neck'' in the softmax, by eliminating the parallel slow-weight path (green bar).
Surprisingly, this improved accuracy slightly over the previous ablation, although the improvement came with increased variance.
Conversely, the performance drops slightly from that of the baseline if we remove the slow weight path from layer $L$.
For both cases of fast weights in the softmax layer, we do not apply the random label projection $R$.
Increasing the number of layers with fast weights did not yield a performance increase, as indicated by the yellow bar.
We also tried truncating the error gradients of the task loss at the fast weights so that no learning signal passes through the Hebbian mechanism. This prevents internal representations from adapting to the learning rule and hurts the performance, as indicated by the violet bar.

We implemented all models in the Chainer framework \cite{chainer_learningsys2015} and tested on an Nvidia Titan X GPU. In Figure~\ref{fig:timing_mini} we see that fwCNN (Hebb) is very efficient during both training and inference. Comparing the runtime of fwCNN (Hebb) and fwCNN ($\nabla$), there is around a 3 ms/task difference from the additional backpropagation and mapping of gradients to fast weights with $g$.
The feedforward procedure takes most of the training time, whereas the description phase during which the learners adapt is slow in inference. MetaNet \cite{pmlr-v70-munkhdalai17a} is the slowest to train and test. The main bottleneck there is the example-level fast weights that must be constructed from memory for each input. Since they modify activations rather than (more numerous) weights, adaCNNs~\cite{munkhdalai2017learning} are also relatively fast.

\section{Conclusion and Future Work}
We developed a model for metalearning that combines slow weights for representation learning with fast weights generated by a Hebbian learning rule for binding labels to representations.
Our model achieves state-of-the-art performance on three one-shot learning benchmarks.

A limitation of our model is that it only incorporates fast weights alongside fully-connected layers.
It is unclear how to incorporate similar associative memory mechanisms into, e.g., LSTMs while still taking advantage of their gates and memory cells.
More generally, we wish to explore one-shot adaptive models in RL or unsupervised learning settings. 

Existing one-shot benchmarks simply relabel fixed underlying class identities from task to task.
A more interesting benchmark might use permuted sets of conjunctions of underlying features to compose flexible, varying classes.



\newpage

\small

\bibliographystyle{plainnat}
\bibliography{example_paper}

\clearpage

\appendix

\section{Additional Implementation Details}
\label{sec:imp_details}

The hyperparameters for our models are listed in Tables \ref{tab:hyper-img} and \ref{tab:hyper-lm}. We did not use dropout in our models. CNN with 32 and 64 filters were trained for Mini-ImageNet and Omniglot benchmarks, respectively. We used LSTMs with 200 units for one-shot and LSTMs with 300 units for two- and three-shot language modelling tasks. We occasionally observed difficulty the LSTM+fwFFN model on the one-shot task and reducing the hidden unit size from 300 to 200 helped.

The neural network weights were initialized using \citet{he2015delving}'s method.
No gradient clipping was performed for optimization. We listed the setup for optimizers in Table \ref{tab:hyper-img} and \ref{tab:hyper-lm}. For Adam optimizer, the rest of the hyperparameters were set to their default values (i.e., $\beta_1 = 0.9$, $\beta_2 = 0.999$, and $\epsilon = 10^{-8}$).

For fwCNN (Hebb) model, we found that adding a noise drawn from a uniform distribution $unif\left(-0.01, 0.01\right)$ before random projection slightly improves the performance. The random projection matrix $R_L$ was initialized from a uniform distriubtion $unif\left(-1.0/\sqrt{d_L}, 1.0/\sqrt{d_L}\right)$.

Although different parameterizations for the function $g$ for mapping gradients to fast weights may improve the performance, for simplicity we used a 3-layer MLP with leaky $\relu$ activation with 40 units per layer. This MLP acts coordinate-wise and maps the gradients to corresponding fast weights independently.

Models were implemented using the Chainer~\citep{chainer_learningsys2015} framework\footnote{https://chainer.org/}.

\begin{table}[h]
  \centering
  \caption{Hyperparameters for few-shot image classification tasks}
  \label{tab:hyper-img}
  \small
  \begin{tabular}{lccccc}
    \toprule
        \bf Model & \bf Layers & \bf Filters & \bf Hidden units in FC layer $d_L$ & \bf Optimizer \\
    \midrule
        fwCNN ($\nabla$) & 5 & 32/64 & 288 &	Adam ($\alpha$=0.001) \\
        fwCNN (Hebb) & 5 &	32/64 & 288 &	Adam ($\alpha$=0.001) \\
        fwResNet ($\nabla$) & 4 &	64, 96, 128, 256 &	1024 & Adam ($\alpha$=0.001) \\
        fwResNet (Hebb) & 4 & 64, 96, 128, 256 & 1024 & Adam ($\alpha$=0.001) \\
    \bottomrule
  \end{tabular}
\end{table}

\begin{table}[h]
\vskip -0.5cm
  \caption{Hyperparameters for few-shot language modelling tasks}
  \label{tab:hyper-lm}
  \small
  \centering
  \begin{tabular}{lcccc}
    \toprule
        \bf Model & \bf Hidden unit size in FC layer $d_L$ & Embedding size & \bf Optimizer \\
    \midrule
        2-layer LSTM + fwFFN ($\nabla$) &	300, 200 & 300 &	Adam ($\alpha$=0.001) \\
        2-layer LSTM + fwFFN (Hebb) &	300, 200 & 300 & Adam ($\alpha$=0.001) \\
    \bottomrule
  \end{tabular}
\end{table}

\end{document}